\documentclass[10pt,twocolumn,letterpaper]{article}

\usepackage{cvpr}              

\usepackage{graphicx}
\usepackage{amsmath}
\usepackage{amssymb}
\usepackage{booktabs}
\usepackage{float}
\usepackage{caption}
\usepackage[pagebackref,breaklinks,colorlinks]{hyperref}
\usepackage{multirow}

\usepackage[capitalize]{cleveref}
\crefname{section}{Sec.}{Secs.}
\Crefname{section}{Section}{Sections}
\Crefname{table}{Table}{Tables}
\crefname{table}{Tab.}{Tabs.}

\DeclareMathOperator\CE{CE}

\begin{document}

\title{Cross-modal Face- and Voice-style Transfer}

\author{Naoya Takahashi \and Mayank K. Singh \and Yuki Mitsufuji \and\\
Sony Group Corporation, Japan\\
{\tt\small \{Naoya.Takahashi, Mayank.A.Singh, Yuhki.Mitsufuji\}@sony.com}
}
\maketitle

\begin{abstract}
Image-to-image translation and voice conversion enable the generation of a new facial image and voice while maintaining some of the semantics such as a pose in an image and linguistic content in audio, respectively. They can aid in the content-creation process in many applications. However, as they are limited to the conversion within each modality,
matching the impression of the generated face and voice remains an open question. 
We propose a cross-modal style transfer framework called XFaVoT that jointly learns four tasks:
image translation and voice conversion tasks with audio or image guidance, which enables the generation of ``face that matches given voice” and ``voice that matches given face”, and intra-modality translation tasks with a single framework. Experimental results on multiple datasets shows that XFaVoT achieves cross-modal style translation of image and voice, outperforming baselines in terms of quality, diversity, and face--voice correspondence.
\end{abstract}


\begin{figure}[t]
  \centering
  \includegraphics[width=\linewidth]{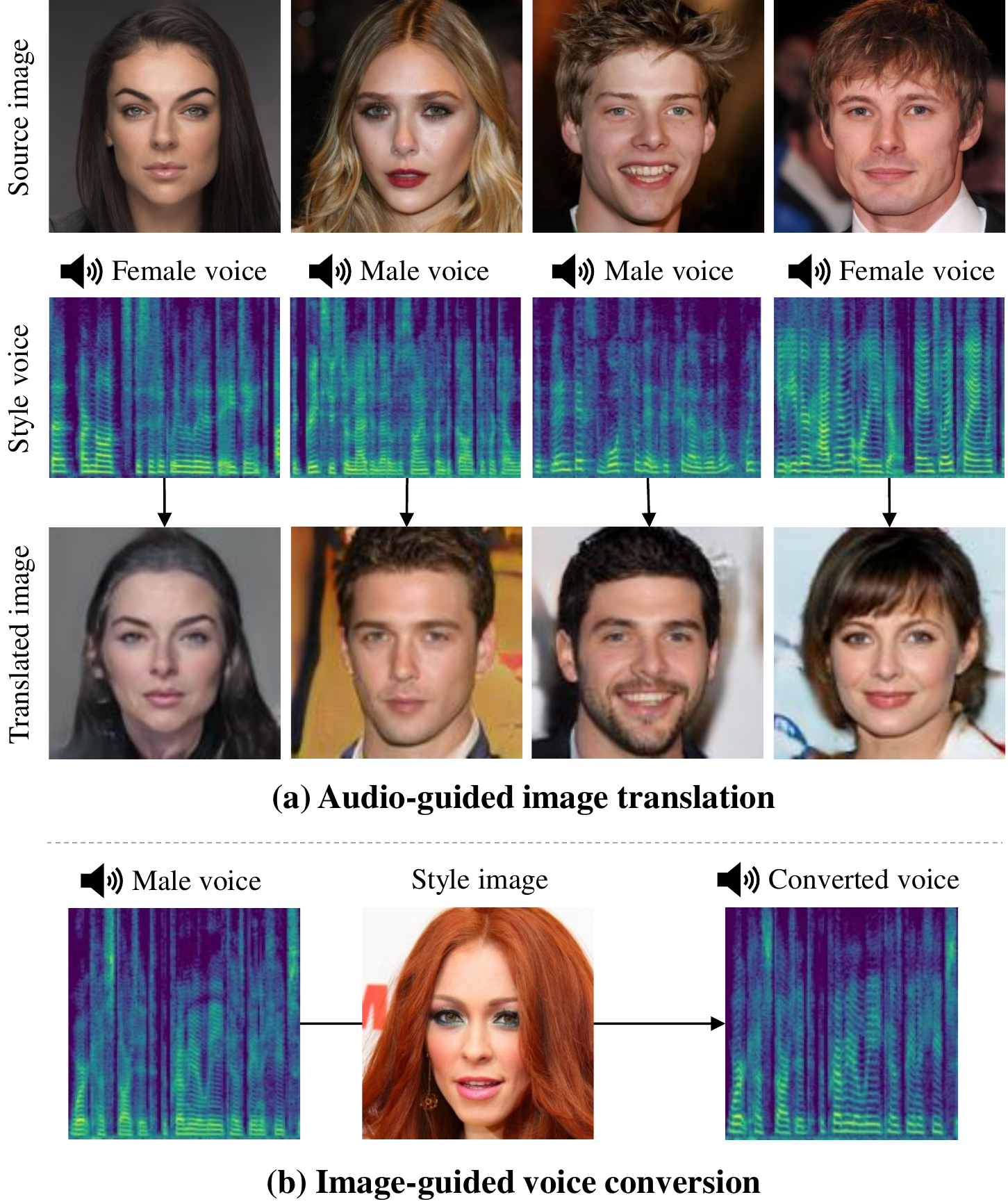}
  \caption{Results of cross-modal style transfer. (a) Audio-guided face synthesis transforms source image style to match given voice character while maintaining pose and contour. (b) Image-guided voice conversion converts source voice to match given style image while maintaining linguistic content. Audio samples are available on our anonymous demo page$^{\ref{fn:demo}}$.}
  \label{fig:resutls}
\end{figure}

\section{Introduction}
\label{sec:intro}
Image-to-image translation \cite{Isola2017I2I} and voice conversion \cite{Kaneko2018UPVC} have been widely studied due to their wide range of applications such as character creation and editing for professional content, social media content, anonymization, and avatars. Reference-guided image-to-image translation enables the modification of specific parts in a human face image, such as hairstyle and color, on the basis of the reference image \cite{Huang2018I2I,Lee2018I2I, Mao2019I2I,Choi20StarGANv2}, while voice conversion generates a new voice that maintains the linguistic content of a source utterance \cite{chou2019adainvc,Lin21FragmentVC,Li21StarGANv2VC}. Although image-to-image translation and voice conversion provide easy and intuitive ways to manipulate the face and voice, they are independently studied in each modality, and their relationships are largely neglected. Therefore, considerable manual effort is required to match the impression of the generated face and voice. 

To address this problem, we propose XFaVoT, a single framework for cross-modal face- and voice-style transfer that enables the generation of ``\textit{a face that matches a user-provided voice}" and ``\textit{a voice that matches a user-provided image}", as shown in \Cref{fig:resutls}. 
XFaVoT jointly learns four tasks in a single framework; audio-guided image translation, image-guided image translation, audio-guided voice conversion, and image-guided voice conversion. 
The proposed model learns style embedding space, which is common to audio and image modalities and consistent with the face and voice from the same speakers. Image and audio generators learn to generate a new face and voice that reflect the style embedding obtained from either a reference face using an image encoder or reference voice using an audio encoder. 
Using face--voice pairs extracted from human talking videos, we train the proposed model on the four tasks while regularizing the consistency of the style vectors extracted from the audio and image of the same speaker via contrastive learning. 
To ensure both generalizability to unseen speakers and accurate modeling of speaker-dependent voice characteristics, we introduce dual-domain discriminators and a mapping network that accept two types of domain codes, \textit{gender} and \textit{speaker-identity} codes.
To address the scarcity of a high-quality clean audio-visual dataset, we further propose leveraging unpaired audio-only and image-only datasets by switching the domain code and loss functions to appropriately incorporate the unpaired data.

Experimental results on the GRID, CelebA-HQ+VCTK, and LRS3+Lip2Wav datasets show that XFaVoT can generate high-quality images and voices that reflect given references in other modalities, outperforming baselines in the audio-guided image translation and image-guided voice conversion tasks without sacrificing the performance of the intra-modal translation tasks.
Samples including audio are available on our demo page \footnote{\url{https://t-naoya.github.io/xfavot/}\label{fn:demo}} and supplemental material. 

\section{Related work}

\textbf{Image-to-image translation.}\hspace{2mm}
Early image-to-image translation studies focus on learning mapping functions between two domains \cite{Isola2017I2I, Zhu2017CycleGAN,Liu2017I2I}. However, they are known to learn a deterministic mapping even with stochastic noise inputs. To improve diversity, several methods have been proposed such as marginal matching \cite{Almahairi2018AugCycleGAN}, latent regression \cite{Zhu2017MMI2I,Huang2018MMI2I}, diversity regularization \cite{Yang2019I2I,Mao2019I2I}, and guidance of reference images \cite{Jung2018I2I,Cho2019I2I,Ma2019I2I,Park2019I2I}. As these methods require separate models for each combination of two domains, the training cost becomes expensive as the number of domains increase. To address the scalability, unified frameworks have been proposed \cite{Choi2018StarGAN,Hui2018I2I,Liu2019I2I,Choi20StarGANv2}. StarGANv2 \cite{Choi20StarGANv2} learns the mappings between all available domains using a single generator. Using domain-specific branches for a discriminator and mapping network, StarGANv2 is shown to generate diverse images in the target domains.
Recently, contrastive learning has been applied to enhance spatial correspondence \cite{Park2020ContrastiveI2I,Zheng2021I2I, Wang2021CI2I,Jung2022I2I}.
Several studies attempt to improve image quality by enforcing consistency in the local structure \cite{Ko2022DCR} and spatial perturbation \cite{Xu2022MSPC}. 
Other studies focus on text-guided image manipulation \cite{Dong2017T2I,Nam2018T2I,Li2020ManiGAN,Patashnik2021StyleCLIP,Xia2021TediGAN}, which aims at translating images on the basis of language. In contrast, our model translates images on the basis of voice, which provides a direct way to plausibly match the face image to the given voice.
\vspace{2mm}\\
\textbf{Voice conversion.}\hspace{2mm}
Early unsupervised voice conversion approaches, which do not require parallel data of different speakers speaking the same content, convert voices between two speakers \cite{Xie2016UPVC,Kinnunen2017UPVC,Kaneko2018UPVC,Kaneko2018WaveCycleGAN}.
To improve scalability, many-to-many voice conversion models have been actively studied \cite{Chou2018M2MVC,Kameoka2018StarGANVC,Kaneko2019StarGANVC2,Takahashi21,Li21StarGANv2VC}. Li et al.\cite{Li21StarGANv2VC} adopt StarGANv2 to many-to-many voice conversion and achieve state-of-the-art performance. However, the target speakers are still limited to those seen during the training. Recently, one-shot voice conversion that enables any-to-any voice conversion has been actively investigated \cite{chou2019adainvc, Wu20VQVCp, Chen21AGAINVC, Lin21FragmentVC}. AdaIN-VC \cite{chou2019adainvc} uses a speaker encoder to extract speaker embeddings and condition the decoder using adaptive instance normalization (AdaIN) layers. Fragment-VC \cite{Lin21FragmentVC} uses a cross-attention mechanism to use fragments from reference samples to produce a converted voice.
Our method can perform one-shot voice conversion in both audio- and image-guided voice conversion modes.
\vspace{2mm}\\
\textbf{Cross-modal audio-visual style transfer.}\hspace{2mm}
The most closely related work in audio-guided image translation is probably sound-guided semantic image manipulation (SGSIM) \cite{Lee2022SGIM}, which first learns an audio encoder that maps audio into the latent space of the contranstive language-image pretraining (CLIP) model then uses the encoder to search a latent space of StyleGAN2 \cite{Karras2020StyleGAN2} to generate an image that has a similar CLIP latent code with that obtained from the audio. 
Li et al. propose learning a sound-guided stylization of landscape images from unlabeled video on a hike using noise contrastive estimation (NCE) \cite{Li2022}. Other studies explore stylizing images on the basis of music \cite{Lee2020Music2Image, Jeong2021Traumerai} by mapping the music into StyleGAN's latent space.

On image-guided voice conversion, we are aware of only one work called cross-modal voice conversion (CVC) \cite{Kameoka2019CMVC} in which a variational autoencoder (VAE)-based voice conversion model is conditioned on a face image to specify the speaker. However, that study is limited to artificial data which randomly combine a face and voice from different datasets on the basis of gender and age attributes; hence, the model fails to learn correspondence other than the attributes. Our model is based on adversarial training and learns the face-voice correspondence on real face-voice pairs while reasonably augmenting data with synthetic face-voice pairs to improve generalization and diversity.  
\vspace{2mm}\\
\textbf{Face--voice correspondence.}\hspace{2mm}
Neurocognitive studies indicate that neuro-cognitive pathways for voices share a common structure with that for faces \cite{Ellis1984Neurocognitive}, and human perception implicitly recognizes the association of faces to voices \cite{Belin2004Neurocognitive}.
Empirical studies have shown the ability of humans to associate voices of unknown individuals to pictures of their faces \cite{Kamachi2003,Mavica2013}. 
A few studies attempt to reconstruct face from voice using a deterministic autoencoder \cite{Oh2019Speech2Face} and generative adversarial network \cite{Wen2019FaceReconst} and show promising results. Different from these works, one of our goals is the image translation with audio guidance which maintains the domain-invariant characteristics (e.g. pose) of the source image while translating the style to match the reference voice.

\begin{figure*}[t]
  \centering
  \includegraphics[width=\linewidth]{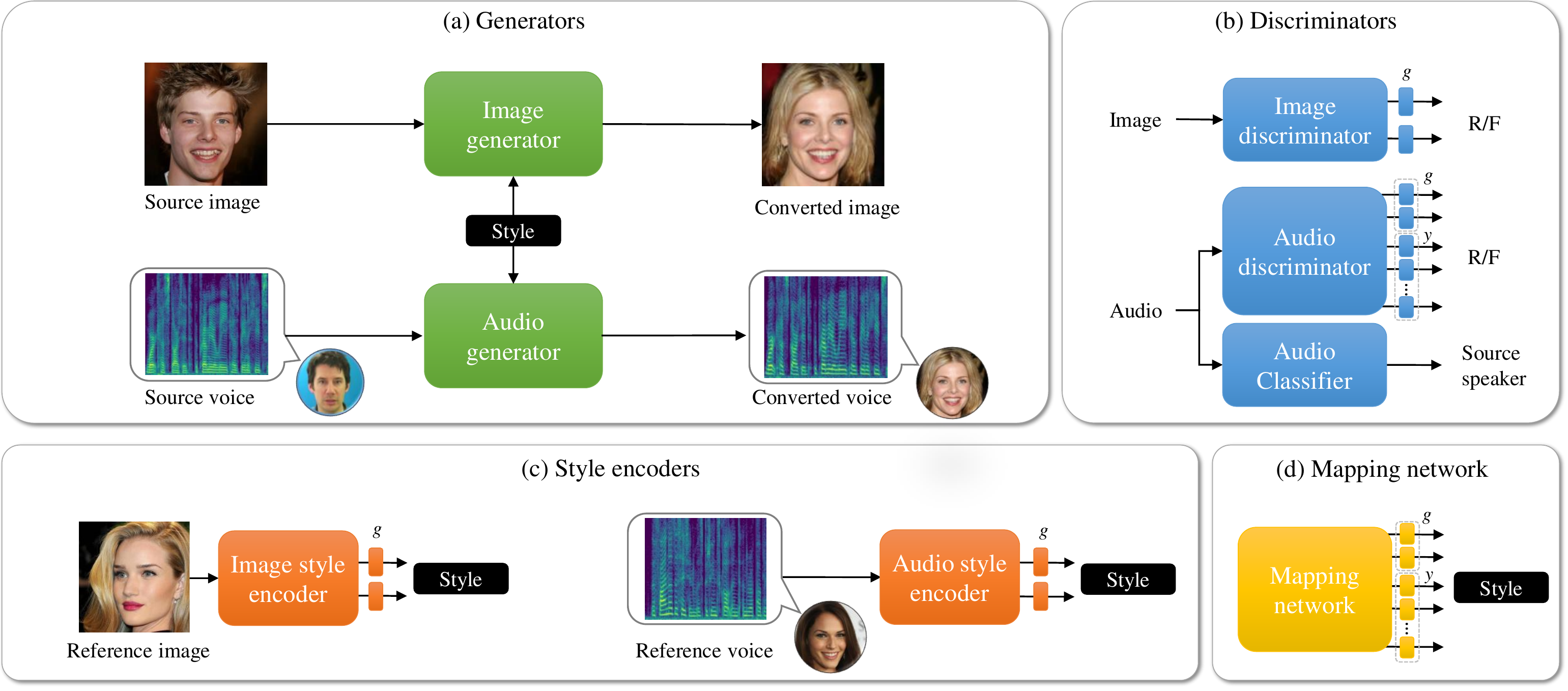}
  \caption{Overview of XFaVoT. (a) The image and audio generators synthesize image and audio that reflect the style vector, respectively. (b) The image and audio discriminators distinguish between real and fake samples from multiple domains via domain-specific heads. Gender code $g$ and speaker code $y$ are used for identifying the domain-specific head. The audio classifier identifies the source speaker of converted voices. (c) Image and audio style encoders extract the style vector from reference image and voice, respectively. The gender code $g$ is used to specify the domain specific head. (d) The mapping network transforms a latent code into the style vectors via domain-specific heads, which accept both $g$ and $y$.}
  \label{fig:overview}
\end{figure*}


\section{Proposed method}
\subsection{Framework}
XFaVoT jointly learns the four tasks, image translation and voice conversion with audio or image guidance. The proposed model is inspired by an image translation model called StarGANv2 \cite{Choi20StarGANv2} and is largely extended to operate on multiple modalities and multiple tasks with dual-domain types. We consider two types of domains, gender $g\in\mathcal{G}=\{male, female\}$ and speaker identity $y\in\mathcal{Y}$, which are common for audio (voice) and image (face) modalities. 
\Cref{fig:overview} shows an overview of XFaVoT, which consists of the following eight modules. \vspace{1mm}\\
\textbf{Generators}\hspace{2mm} The image generator $G^{im}$ translates a source image $\mathbf{x}$ into output image $G^{im}(\mathbf{x},\mathbf{s})$ by reflecting a style provided by the style vector $\mathbf{s}$, which is provided by either the image encoder $E^{im}$, audio encoder $E^{au}$ or mapping network $F_{g,y}$. When the $\mathbf{s}$ is produced by the $E^{im}$, the task is referred to as image-guided image translation, while the $E^{au}$ is used to produce the $\mathbf{s}$ for the audio-guided image translation.  
Similarly, the audio generator $G^{au}$ takes as input a mel-spectrogram of the source voice $\mathbf{a}$ and outputs a mel-spectrogram $G^{au}(\mathbf{a},\mathbf{s})$ reflecting the style vector. When the style vector produced by the $E^{au}$ is used, the task results in one-shot voice conversion, which we refer to as audio-guided voice conversion for consistency. When the style vector produced by the $E^{im}$ is used, we refer to the task as image-guided voice conversion. \vspace{1mm}\\
\textbf{Style encoders}\hspace{2mm} The image style encoder $E^{im}_g$ and audio style encoder $E^{au}_g$ extract the style vector from a reference image and mel-spectrogram of a reference audio in domain $g$, respectively. Both encoders have gender-domain-specific heads upon the main networks that is common to all domains. \vspace{1mm}\\
\textbf{Mapping network}\hspace{2mm} Given a latent code $\mathbf{z}\in\mathcal{Z}$ sampled from a prior distribution, the mapping network $F_{g,y}$ transforms $\mathbf{z}$ into the style vector $\mathbf{s}=F_{g,y}(\mathbf{z})$. We use a single mapping network, which promotes the consistency of the style vectors between the audio and image modalities. 
The mapping network consists of a multi-layer perceptron with domain-specific heads for both $g\in\mathcal{G}$ and $y\in\mathcal{Y}$. \vspace{1mm}\\
\textbf{Discriminators}\hspace{2mm} The image discriminator $D^{im}_g$ consists of a common network followed by domain-specific binary classification heads that distinguish whether the input is a real image of domain $g$ or a fake image translated by the $G^{im}$. In contrast, the audio discriminator $D^{au}_{g,y}$ takes as an input the mel-spectrogram and distinguishes whether it is a real or fake one produced by the $G^{au}(\mathbf{a},\mathbf{s})$. The $D^{au}_{g,y}$ has the same architecture as the $D^{im}_g$ except for the domain-specific heads, where the heads specific for $g\in\mathcal{G}$ and $y\in\mathcal{Y}$ are both available.
We also introduce the audio classifier $C$ that predicts the source speaker of the converted voice $G^{au}(\mathbf{a},\mathbf{s})$. 

\subsection{Training objectives}
The aim of XFaVoT is to learn two mapping functions $G^{im}:\mathcal{X}_{g}\rightarrow \mathcal{X}_{\tilde{g}}$ that converts an image $\mathbf{x}\in\mathcal{X}_{g}$ from the source domain $g\in\mathcal{G}$ to a sample $\hat{\mathbf{x}}\in\mathcal{X}_{\tilde{g}}$ in the target domain  $\tilde{g}\in\mathcal{G}$ and $G^{au}:\mathcal{A}_{g,y}\rightarrow \mathcal{A}_{\tilde{g},\tilde{y}}$  that converts a mel-spectrogram of source voice $\mathbf{a}\in\mathcal{A}_{g,y}$ from source domain $g\in\mathcal{G}, y\in\mathcal{Y}$ to a sample $\hat{\mathbf{a}}\in\mathcal{A}_{\tilde{g},\tilde{y}}$ in the target domain  $\tilde{g}\in\mathcal{G},\tilde{y}\in\mathcal{Y}$.
The speaker domain codes $y,\tilde{y}\in\mathcal{Y}$ can be unknown for some of the data, as discussed in \cref{sec:dataset}.
We jointly train the eight modules using audio and image data for audio- and image-guided image translation tasks and audio- and image-guided voice conversion tasks. We sample reference-domain codes $\tilde{g}$, $\tilde{y}$ and a style vector $\mathbf{s}$ via either the image style encoder $E^{im}_{\tilde{g}}$, audio style encoder $E^{au}_{\tilde{g}}$, or mapping network $F_{\tilde{g},\tilde{y}}$ and train the model using the following loss functions. \vspace{2mm}\\
\textbf{Adversarial loss}\hspace{2mm}
The audio and image generators take an input image $\mathbf{x}$ and mel-spectrogram $\mathbf{a}$ along with a style vector $\mathbf{s}$ and learn to generate a new image $G^{im}(\mathbf{x},\mathbf{s})$ and mel-spectrogram $G^{au}(\mathbf{a},\mathbf{s})$ via adversarial loss, respectively;
\begin{equation}
\begin{split}
 &L_{adv}=\\
 &\mathbb{E}_{\mathbf{x},g}[\log D^{im}_{g}(\mathbf{x}))]+
 \mathbb{E}_{\mathbf{x},\tilde{g},\mathbf{s}}[\log (1-D^{im}_{\tilde{g}}(G^{im}(\mathbf{x},\mathbf{s})))] +\\
&\mathbb{E}_{\mathbf{a},g,y}[\log D^{au}_{g,y}(\mathbf{x}))]+
 \mathbb{E}_{\mathbf{a},\tilde{g},\tilde{y},\mathbf{s}}[\log (1-D^{au}_{\tilde{g},\tilde{y}}(G^{au}(\mathbf{a},\mathbf{s})))],
\end{split}
\end{equation}
where $\tilde{g}$ and $\tilde{y}$ are the gender and speaker identity codes of the reference speaker which corresponds to the style vector. During training, we randomly choose either gender- or speaker-specific heads to compute $D_{g,y}^{au}$ 
\vspace{2mm}\\
\textbf{Style reconstruction loss}\hspace{2mm} The style reconstruction loss is used
to ensure that the style code can be reconstructed from the generated samples.
\begin{equation}
\begin{split}
 L_{sty}=&\mathbb{E}_{\mathbf{x},\tilde{g},\mathbf{s}}[||\mathbf{s}-E^{im}_{\tilde{g}}(G^{im}(\mathbf{x},\mathbf{s}))||_1] +\\
&\mathbb{E}_{\mathbf{a},\tilde{g},\mathbf{s}}[||\mathbf{s}-E^{au}_{\tilde{g}}(G^{au}(\mathbf{a},\mathbf{s}))||_1]
\end{split}
\end{equation}
\vspace{2mm}\\
\textbf{Style diversification loss}\hspace{2mm}
To enable the generators to produce diverse images and audio, we employ the diversity sensitive loss \cite{Mao2019,Yang2019I2I},
\begin{equation}
\begin{split}
 &L_{ds}=\\
 &\mathbb{E}_{\mathbf{x},\tilde{g},\mathbf{s_1},\mathbf{s_2}}[||E^{im}_{\tilde{g}}(G^{im}(\mathbf{x},\mathbf{s_1}))-E^{im}_{\tilde{g}}(G^{im}(\mathbf{x},\mathbf{s_2}))||_1] +\\
 &\mathbb{E}_{\mathbf{a},\tilde{g},\mathbf{s_1},\mathbf{s_2}}[||E^{au}_{\tilde{g}}(G^{au}(\mathbf{a},\mathbf{s_1}))-E^{au}_{\tilde{g}}(G^{au}(\mathbf{a},\mathbf{s_2}))||_1],
\end{split}
\end{equation}
where $\mathbf{s_1},\mathbf{s_2}\in\mathcal{S}_{\tilde{g},\tilde{y}}$ are two randomly sampled style vectors from domain $\tilde{g},\tilde{y}$. \vspace{2mm}\\
\textbf{Cycle consistency loss}\hspace{2mm}
To preserve the domain-invariant characteristics (e.g. pose in image and linguistic content in voice), we employ the cycle consistency loss \cite{Choi2018StarGAN}
\begin{equation}
\begin{split}
 L_{cyc}=&\mathbb{E}_{\mathbf{x},g, \tilde{g},\mathbf{s}}[||\mathbf{x}-G^{im}_{\tilde{g}}(G^{im}(\mathbf{x},\tilde{\mathbf{s}}), \hat{\mathbf{s}}^{im})||_1] +\\
 &\mathbb{E}_{\mathbf{x},g, \tilde{g},\mathbf{s}}[||\mathbf{a}-G^{au}_{\tilde{g}}(G^{au}(\mathbf{a},\tilde{\mathbf{s}}), \hat{\mathbf{s}}^{au})||_1],
\end{split}
\end{equation}
where $\hat{\mathbf{s}}^{im}=E^{im}_g(\mathbf{x})$ and $\hat{\mathbf{s}}^{au}=E^{au}_g(\mathbf{a})$ are the style vectors of the source image $\mathbf{x}$ and audio $\mathbf{a}$, respectively. \vspace{1mm}\\
\textbf{Audio auxiliary losses}\hspace{2mm}
Voice conversion requires maintaining the linguistic content of the source utterance while converting the voice character. To enable this, we further introduce auxiliary losses, as suggested in \cite{Li21StarGANv2VC}. We use the speech consistency loss using a pretrained automatic speech recognition model $A$ as $L_{asr}=\mathbb{E}_{\mathbf{a},\mathbf{s}}[||A(\mathbf{a})-A(G^{au}(\mathbf{a},\mathbf{s}))||_1]$.
To facilitate learning and promote consistency in pitch and rhythm, we further introduce pitch consistency loss $L_{F0}=\mathbb{E}_{\mathbf{a},\mathbf{s}}||\bar{\mathcal{F}}(\mathbf{a})-\bar{\mathcal{F}}(G^{au}(\mathbf{a},\mathbf{s}))||_1$ and the norm consistency loss $L_{norm}=\mathbb{E}_{\mathbf{a},\mathbf{s}}||N(\mathbf{a})-N(G^{au}(\mathbf{a},\mathbf{s}))||_1$, where $\bar{\mathcal{F}}(\mathbf{a})= \mathcal{F}(\mathbf{a})/||\mathcal{F}(\mathbf{a})||_1$ denotes the normalized fundamental frequency (F0) obtained using a pretrained F0 estimation network $\mathcal{F}$, and $N(\cdot)$ is the frame-wise energy. To further facilitate speaker specific information and promote conversion, we use the adversarial classification loss. The audio source classifier $C$ is trained to identify the source speaker via classification loss $\mathcal{L}_{cls}=\mathbb{E}_{\mathbf{a},\mathbf{s}}[\CE(C(G^{au}(\mathbf{a},\mathbf{s})), y)]$, and the audio generator is trained to fool the classifier via the adversarial classification loss $\mathcal{L}_{adcl}=\mathbb{E}_{\mathbf{a},\mathbf{s}}[\CE(C(G^{au}(\mathbf{a},\mathbf{s})), \tilde{y})]$.
where $\CE$ denotes the cross-entropy loss, $y$ the source speaker, and $\tilde{y}$ the reference speaker.\vspace{2mm}\\
\textbf{Cross-modality style consistency}\hspace{2mm}
Using the common mapping network for image and audio modalities promotes the consistency of style vectors obtained from the image and audio encoders. However, it may not be sufficient to achieve speaker-identity-level consistency as the image encoder and discriminator use only the gender domain code $g$. The style vectors obtained from the encoders with the face image and voice from the same speaker may not be close to each other. To further ensure consistency across the modalities,  we use the infoNCE loss \cite{Alayrac2020infoNCE}. Assuming audio-image pair data, we sample pairs $(\mathbf{a}_i,\mathbf{x}_i), i=1,\dots,N$ from N speakers and compute style vectors $(\mathbf{s}^{au}_i,\mathbf{s}^{im}_i)=(E^{au}_g(\mathbf{a}_i),E^{im}_g(\mathbf{x}_i))$. We then compute the following loss function for the $i$th pair 
\begin{equation}
 l_i=-\log \frac{\exp(\langle s^{au}_i, s^{im}_i\rangle/\tau)}{\sum_{j=1}^N{\exp(\langle s^{au}_i, s^{im}_j\rangle/\tau)}},
\end{equation}
where $\langle\cdot,\cdot\rangle$ denotes the cosine similarity, and $\tau$ is a temperature parameter. The cross-modality style consistency loss is obtained by the average of the loss function $L_{csc}=\frac{1}{N}\sum_{i=1}^Nl_i$.\vspace{2mm}\\
\textbf{Full objectives}\hspace{2mm} Our full objective function for generators can be
summarized as follows:
\begin{equation}
\begin{split}
  \min_{G,E,F}&\mathcal{L}_{adv}+\lambda_{sty}\mathcal{L}_{sty}-\lambda_{ds}\mathcal{L}_{ds}+\\
 &\lambda_{cyc}\mathcal{L}_{cyc}+\lambda_{asr}\mathcal{L}_{asr}+\lambda_{F0}\mathcal{L}_{F0}+\\
 &\lambda_{norm}\mathcal{L}_{norm}+\lambda_{adcl}\mathcal{L}_{adcl}+\lambda_{csc}\mathcal{L}_{csc}
\end{split}
\end{equation}
where the generator $G$ and style encoder $E$ include both audio and image modules, and $\lambda_{sty}$, $\lambda_{ds}$, $\lambda_{cyc}$, $\lambda_{asr}$, $\lambda_{F0}$, $\lambda_{norm}$, $\lambda_{advcls}$ and $\lambda_{csc}$ are the hyper parameters for each term.

Our full objective for discriminators is given as:
\begin{equation}
\begin{split}
\min_{D,C}&-\mathcal{L}_{adv}+\lambda_{cls}\mathcal{L}_{cls}
\end{split}
\end{equation}

\subsection{Leveraging dataset in single-modality}
\label{sec:dataset}
To achieve high-fidelity generation, datasets that have high-quality image and audio are required. There are few audio-visual datasets \cite{Cookea2006GRID, Harte2015TCDTIMIT, Prajwal_2020Lip2Wav} that provide videos of human talking with clearly visible face and clean audio without noise and reverberation. However, as such clean data require a controlled environment for recording or careful curation process, dataset size is limited, which hinders the learning of diverse audio and image generation. Although large-scale video datasets collected from the Internet, such as LRS3 \cite{Afouras2018LRS3}, offer diverse videos of people speaking, faces are often blurry and in low-resolution, and the audio contains noise and reverberation. Training the model on such a distorted, unclean dataset hinders the learning of high-quality audio and image generation. 
To address this problem, we propose using high-quality datasets independently available in audio and image domains along with clean (possibly small) audio-visual datasets. Specifically, we use CelebA-HQ \cite{Karras2018PGGAN} for image and VCTK \cite{Yamagishi2019VCTK} for audio. As only the cross-modality style consistency loss $\mathcal{L}_{csc}$ requires audio-visual pair data, we omit $\mathcal{L}_{csc}$ from the full objective when we sample the style vector using an image from the image-only dataset $x^{im}$ and audio from the audio-only dataset $a^{au}$. Since the image-only dataset does not have real audio that corresponds to the image  $x^{im}$, the speaker-identity-specific heads of the $D^{au}$ cannot be trained with the style vector produced using $x^{im}$. In this case, we only train the gender-specific heads. (Hence, we do not require speaker label $y$ for image-only data.) The combination of the style vector source and the use of the loss functions is summarized in \Cref{tab:loss}.

\begin{table}
    \centering
    \small
    \caption{Combination of the style vector, cross-modality style consistency loss $\mathcal{L}_{csc}$, and trainable heads of the audio discriminator $D^{au}$. $\mathbf{a}^{av}$ and $\mathbf{x}^{av}$ denote audio and image from audio-visual dataset, respectively. }
    \begin{tabular}{l|cc}
        \toprule
        Style vector & $\mathcal{L}_{csc}$ &$D^{au}$\\
        \midrule
        $E^{au}_g(\mathbf{a}^{av})$ & \checkmark & $g,y$\\
        $E^{im}_g(\mathbf{x}^{av})$ & \checkmark & $g,y$\\
        $E^{au}_g(\mathbf{a}^{au})$ &  & $g,y$\\
        $E^{au}_g(\mathbf{x}^{im})$ &  & $g$\\
        \bottomrule
    \end{tabular}

    \label{tab:loss}
\end{table}

\subsection{Implementation details}
We base our model implementation on the offical code of StarGANv2 \cite{Choi20StarGANv2}\footnote{\url{https://github.com/clovaai/stargan-v2}} and use the same network architecture for the encoders, discriminators, and mapping network. The audio classifier has the same architecture as the discriminator. For audio representation, we use 80-band mel-spectrogram with an fast-Fourier-transform size of 2048 and hop size of 300. The generated mel-spectrograms are converted to wavefrom using the Parallel WaveGAN vocoder \cite{Yamamoto20PWG}.  We provide further details in the Appendix.

\begin{table*}
    \centering
    \small
    \caption{Results of image translation on CelebA-HQ, GRID, and combination of LRS3 and Lip2Wav. For audio-guided image translation on CelebA-HQ, we use audio from VCTK.}
    \label{tab:IT}
    \begin{tabular}{c|l|cc|cc|cc}
        \toprule
        \multirow{2}{*}[-1pt]{Guidance} & \multirow{2}{*}[-1pt]{Model} &\multicolumn{2}{c|}{CelebA-HQ (+VCTK)} &\multicolumn{2}{c|}{GRID} &\multicolumn{2}{c}{LRS3+Lip2Wav} \\
        & &FID [$\downarrow$] &LPIPS [$\uparrow$]&FID [$\downarrow$] &LPIPS [$\uparrow$] &FID [$\downarrow$] &LPIPS [$\uparrow$]\\
        \midrule
        \midrule
        \multirow{2}{*}{Image} & StarGANv2 \cite{Choi20StarGANv2} &	27.9&	\textbf{0.254}&	\textbf{34.8}&	0.179&	72.2&	\textbf{0.266}\\
        &  XFaVoT (Ours) &\textbf{20.6}	&0.169 &	48.0&	\textbf{0.181}&	\textbf{60.4}&	0.248\\
        \midrule
        \multirow{2}{*}{Audio} & SGSIM \cite{Lee2022SGIM} &	148.4&	0.146&	121.4&	0.129&	161.4&	0.120\\
        & XFaVoT (Ours) &	\textbf{38.3}&	\textbf{0.153}&	\textbf{28.3}&	\textbf{0.158}&	\textbf{74.1}&	\textbf{0.238}\\
        \bottomrule
    \end{tabular}
\end{table*}


\section{Experiments}
Our goal is to perform the cross-modal style transfer tasks (audio-guided image translation and image-guided voice conversion) while maintaining the performance of the base model \cite{Choi20StarGANv2} in the image-guided image translation task and extending StarGANv2VC \cite{Li21StarGANv2VC} to the one-shot (audio-guided) voice conversion task. Thus, we evaluate the proposed method on the four tasks.
\vspace{2mm}\\
\textbf{Datasets.}\hspace{2mm} For audio-visual data, we use three datasets, GRID \cite{Cookea2006GRID}, Lip2Wav \cite{Prajwal_2020Lip2Wav}, and LRS3 \cite{Afouras2018LRS3}. As the videos in LRS3 are TED-talk recordings online, their audio is mostly noisy. We manually choose 16 videos that contain relatively less noise and reverberation. The number of speakers in GRID, Lip2Wav, and LRS3 are 33, 4, and 16, respectively. We exclude four speakers from GRID for evaluating the models on unseen speakers. Face images are extracted from video frames and aligned as done in \cite{Karras2018PGGAN}. 
We use CelebA-HQ \cite{Karras2018PGGAN} as the image-only dataset. For audio-only data, VCTK \cite{Yamagishi2019VCTK}, in which utterances from 109 speakers are available, is used. We exclude 30 speakers for evaluation. We resize images to 128$\times$128 resolution and resample audio to 24kHz. The length of audio varies from 6 to 9 s depending on the utterance. The datasets are randomly split into 90 and 10\% for training and validation sets, respectively, except CelebA-HQ, from which we extract 1000 male and female images for the validation set, as done in \cite{Choi20StarGANv2}.\vspace{2mm}\\
\textbf{Baselines.}\hspace{2mm}
To the best of our knowledge, there have been no studies investigating the four translation tasks in audio and image domain interchangeably. Therefore, we consider baselines for each task. For the image-guided image translation task, we use StarGANv2 \cite{Choi20StarGANv2} since our model shares the same architecture in this task. For the audio-guided image translation task, we use SGSIM \cite{Lee2022SGIM}
by following the official implementation
\cite{Lee2022SGIMgit}.
For the audio-guided voice conversion task, we use two state-of-the-art one-shot voice conversion approaches, namely, AdaIN-VC \cite{chou2019adainvc} and Fragment-VC \cite{Lin21FragmentVC}. Finally, we consider CVC \cite{Kameoka2019CMVC} for the image-guided voice conversion task. However, when we train the original CVC model on our dataset, we obtained unsatisfactory results, possibly due to the fact that our datasets are more complex and diverse than the artificial dataset used in the original study \cite{Kameoka2019CMVC}. After exploration, we found that replacing the network architecture with that used in AdaIN-VC \cite{chou2019adainvc} improved voice quality; thus, we use this model as a baseline and refer to as AdaIN-CVC. More details can be found in the supplementary material.\vspace{2mm}\\
\textbf{Evaluation metrics.}\hspace{2mm}
We evaluate the visual quality and diversity of generated images using Frech\'{e}t inception distance (FID) \cite{Heusel2017FID} and learned perceptual image patch similarity (LPIPS) \cite{Zhang2018LPIPS}. Following the evaluation protocol in \cite{Choi20StarGANv2}, we compute FID and LPIPS for the pairs of gender domains (\textit{female}$\leftrightarrows$\textit{male}) within a dataset and report their average values. 
We translate each test image from a source domain into a target domain using ten reference images or voices randomly sampled from the test set of a target domain. 
(The details on evaluation metrics and protocols are further described in supplementary material.)
To evaluate generated voices, we use three metrics; NISQA \cite{Mittag2021NISQA}, speaker similarity (SpkSim), and word error rate (WER). NISQA is a neural network model that predicts the overall mean opinion score of the naturalness of generated speech. SpkSim is the cosine similarity of d-vectors extracted from a speaker verification model
\cite{Resemblyzer},
which is commonly used to evaluate the speaker similarity of samples generated using voice conversion systems. The WER is obtained using the end-to-end speech recognition system \cite{Kim17HybridASR} provided in ESPNet \cite{espnet}. 

\begin{table*}
    \centering
    \small
    \caption{Results of voice conversion on GRID and VCTK. $^*$ indicates the average SpkSim calculated by comparing the output with 10 random utterances of the target speaker used for the reference image. SpkSim scores are not computable on VCTK+CelebA-HQ as there is no ground truth audio for the images in CelebA-HQ.}
    \begin{tabular}{c|l|ccc|ccc}
        \toprule
        \multirow{2}{*}[-1pt]{Guidance} &\multirow{2}{*}[-1pt]{Model} &\multicolumn{3}{c|}{GRID} &\multicolumn{3}{c}{VCTK (+CelebA-HQ)}\\
        & &NISQA [$\uparrow$] &SpkSim [$\uparrow$]&WER [$\downarrow$] &NISQA [$\uparrow$] &SpkSim [$\uparrow$] &WER [$\downarrow$]\\
        \midrule
        \midrule
        - & Ground Truth &4.61 &0.82 &0.30	&4.29 &	0.93 & 0.08 \\	
        \midrule
        \multirow{3}{*}{Audio} & AdaIN-VC\cite{chou2019adainvc} &3.75 &0.75 &0.65	&2.99 &0.78 & 0.31 \\
        & Fragment-VC \cite{Lin21FragmentVC} &3.49 &	0.70 &0.66 &3.00 &0.71 & 0.27\\
        & XFaVoT (Ours) &\textbf{4.65} & \textbf{0.75} &\textbf{0.46} &\textbf{4.56} & \textbf{0.83} & \textbf{0.16} \\
        \midrule
        \multirow{2}{*}{Image} & AdaIN-CVC \cite{Kameoka2019CMVC,chou2019adainvc}&3.45 &0.61$^*$	&0.54 &2.63 &- & 0.21  \\
        & XFaVoT (Ours) &\textbf{4.58} &\textbf{0.66}$^*$ &\textbf{0.48} &\textbf{4.51} &- & \textbf{0.19} \\
        \bottomrule
    \end{tabular}

    \label{tab:VC}
\end{table*}

\subsection{Image translation}
We first evaluate the proposed model on the image- and audio-guided image translation tasks.
\Cref{tab:IT} summarizes the results on CelebA-HQ, GRID, and the combination of LRS3 and Lip2Wav. In the image-guided task, XFaVoT achieves competitive results to StarGANv2. Note that we do not aim at improving the image-guided image translation performance of StarGANv2, but rather achieving audio-guided image translation with competitive performance as the image-guided task. 
Among the different datasets, both models obtain the lowest FID score on CelebA-HQ and the highest score on LRS3+Lip2Wav. This tendency is consistent with the image quality of the datasets; CelebA-HQ provides the highest quality while LRS3+Lip2Wav provides the lowest. 

\begin{figure}[t]
  \centering
  \includegraphics[width=\linewidth]{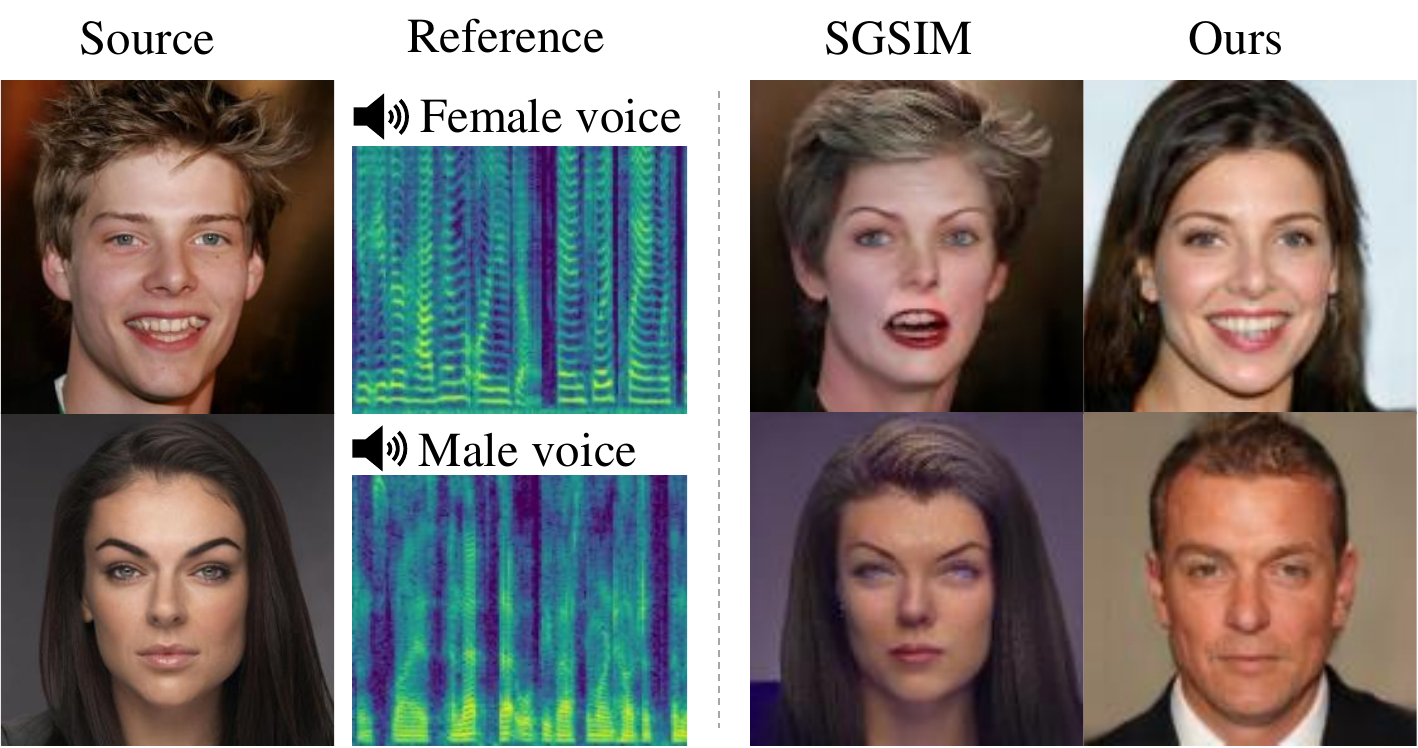}
  \caption{Audio-guided image translation results of SGSIM \cite{Lee2022SGIM} and XFaVoT (ours) on CelebA-HQ. Audio references are sampled from VCTK.}
  \label{fig:sgit}
\end{figure}

\begin{figure}[t]
  \centering
  \includegraphics[width=\linewidth]{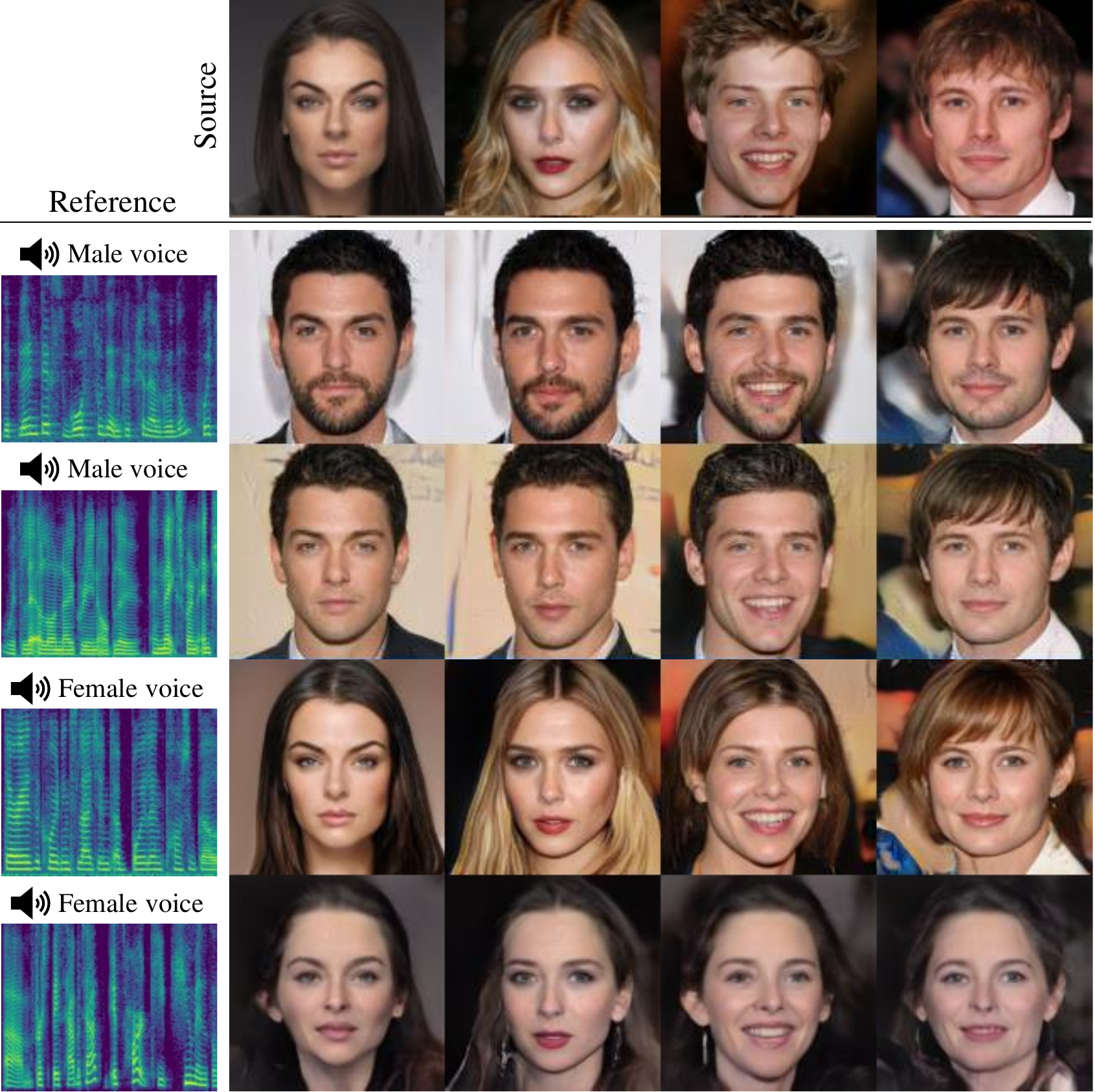}
  \caption{Audio-guided image translation results on CelebA-HQ. Voices are from VCTK, LRS3, and Lip2Wav.}
  \label{fig:samples}
\end{figure}

In the audio-guided image translation task, XFaVoT significantly outperforms SGSIM and achieves competitive results as the image-guided case. This suggests that XFaVoT successfully learns intra- and cross-modal image style translation without compromising the state-of-the-art image generation quality of StarGANv2. \Cref{fig:sgit} shows that XFaVoT can generate more natural images by reflecting the style of voice including gender and age. The styles given by the voice are consistent across difference source images, as shown in \Cref{fig:samples}. Samples with audio and an ablation study can be found in the supplemental material.

\subsection{Voice conversion}
Next, we evaluate the proposed method on the two voice conversion tasks. For the audio-guided voice conversion task, we generate 140 voices using randomly sampled source and reference audio pairs from GRID and VCTK. We omit the evaluation on LRS3 and Lip2Wav due to the lack of transcriptions required to calculate the WER; however, perceptual quality is very similar to the results on the other datasets.
The results are shown in \Cref{tab:VC}. 
XFaVoT outperforms the one-shot voice conversion baselines for all metrics on both datasets, indicating the effectiveness of the StarGANv2-based framework extended for one-shot voice conversion. 

For the image-guided voice conversion task, we generate 180 voices using randomly sampled source audio and reference images from GRID. We also evaluate the models on VCTK, which contains only audio data, by using randomly sampled images from CelebA-HQ as reference. XFaVoT outperforms the AdaIN-CVC baseline on both datasets. 
We encourage readers to refer to our anonymous demo page$^{\ref{fn:demo}}$ for audio samples.

\subsection{Face--voice correspondence}
In this section, we evaluate how well the generated voices and face images correspond to the reference on cross-modal translation tasks, namely, audio-guided image translation and image-guided voice conversion. As we aim to evaluate not only high-level correspondence such as gender but also subjective impressions (e.g. age, physical constitution, energetic/cool, etc.) without access to these labels, we conducted a subjective evaluation.
We create triplets ($x^A,x^B,a$) consisting of two face images $x^A,x^B$ and a voice $a$ and ask evaluators which of the $x^A$ or $x^B$ corresponds to the voice. For audio-guided image translation, we generate two images $\hat{x}_p$ and $\hat{x}_n$ for each source image $x^i$ using two reference voices $a_p, a_n$ of different speakers and create the triplets ($\hat{x}_p,\hat{x}_n,a_p$). For image-guided voice conversion, we generate two converted voices $\hat{a}_p,\hat{a}_n$ for each source voice using two images $x_p,x_n$ of different speakers and create the triplets ($x_p,x_n,\hat{a}_p$). The two images of each triplets are randomly shuffled and shown to evaluators. Hence, the ratio of evaluators choosing the image $\hat{x}_p$ and $x_p$, which we call the correspondence preference ratio (CPR), indicates how well the image and voice correspond to each other. As an upper baseline, we also evaluate ground truth samples by creating the triplets ($x_p,x_n,a$) using an image $x_p$ and voice $a_p$ from the same speaker as positive samples and a randomly sampled image $x_n$ from a different speaker as the negative sample. We create 30 triplets for each model and each task, and 23 evaluators participate in the test.

\begin{figure}[t]
  \centering
  \includegraphics[width=\linewidth]{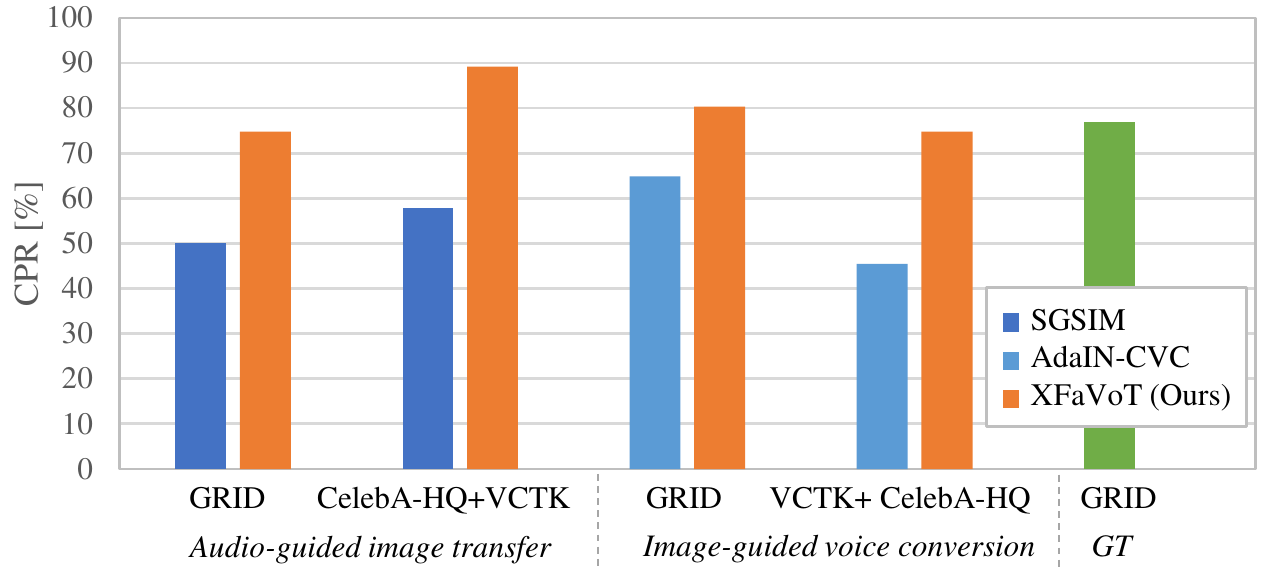}
  \caption{Subjective test results on CelebHQ-VCTK and GRID. CPRs indicate how often the evaluators chose the corresponding pairs used for the generation against random samples. 
  }
  \label{fig:sbj}
\end{figure}

\Cref{fig:sbj} shows the CPRs on GRID and CelebA-HQ+VCTK. XFaVoT outperforms baselines on both audio-guided image translation and image-guided voice conversion tasks. The CPR of the proposed method is close to that of ground truth (GT) on the GRID dataset, suggesting that the face-voice correspondence of generated images and voices is sufficiently high (There is no ground truth for the VCTK+CelebA-HQ dataset as CelebA-HQ and VCTK are independent datasets on image and audio modality, respectively).  
The results on the CelebA-HQ+VCTK dataset highlight how well the model generalizes to the combination of datasets in different modalities since the cross-modality style consistency loss $\mathcal{L}_{csc}$ is not applicable on this combination during the training. Our method achieves a high CPR even in this setting.

\subsection{Latent-guided face and voice generation}
\begin{figure}[t]
  \centering
  \includegraphics[width=\linewidth]{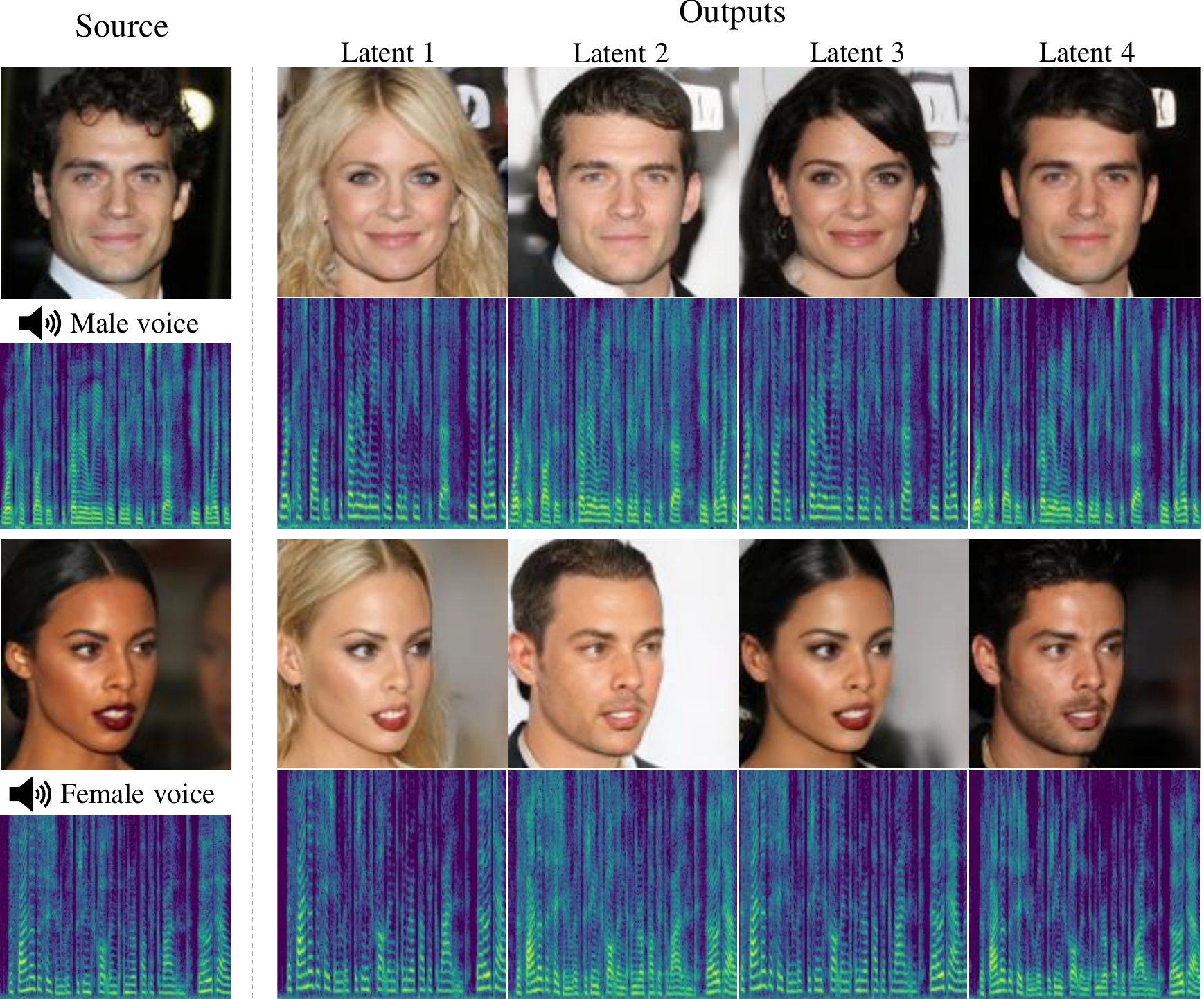}
  \caption{Image translation and voice conversion results using the mapping network with four different latent codes. }
  \label{fig:map}
\end{figure}

As XFaVoT learns the mapping network that maps latent code to the style vector, we can sample a face and voice style from the latent distribution and perform latent-guided face translation and voice conversion. As illustrated in \Cref{fig:map}, the proposed model generates diverse faces and voices without reference from the source face and voice.


\begin{figure}[t]
  \centering
  \includegraphics[width=\linewidth]{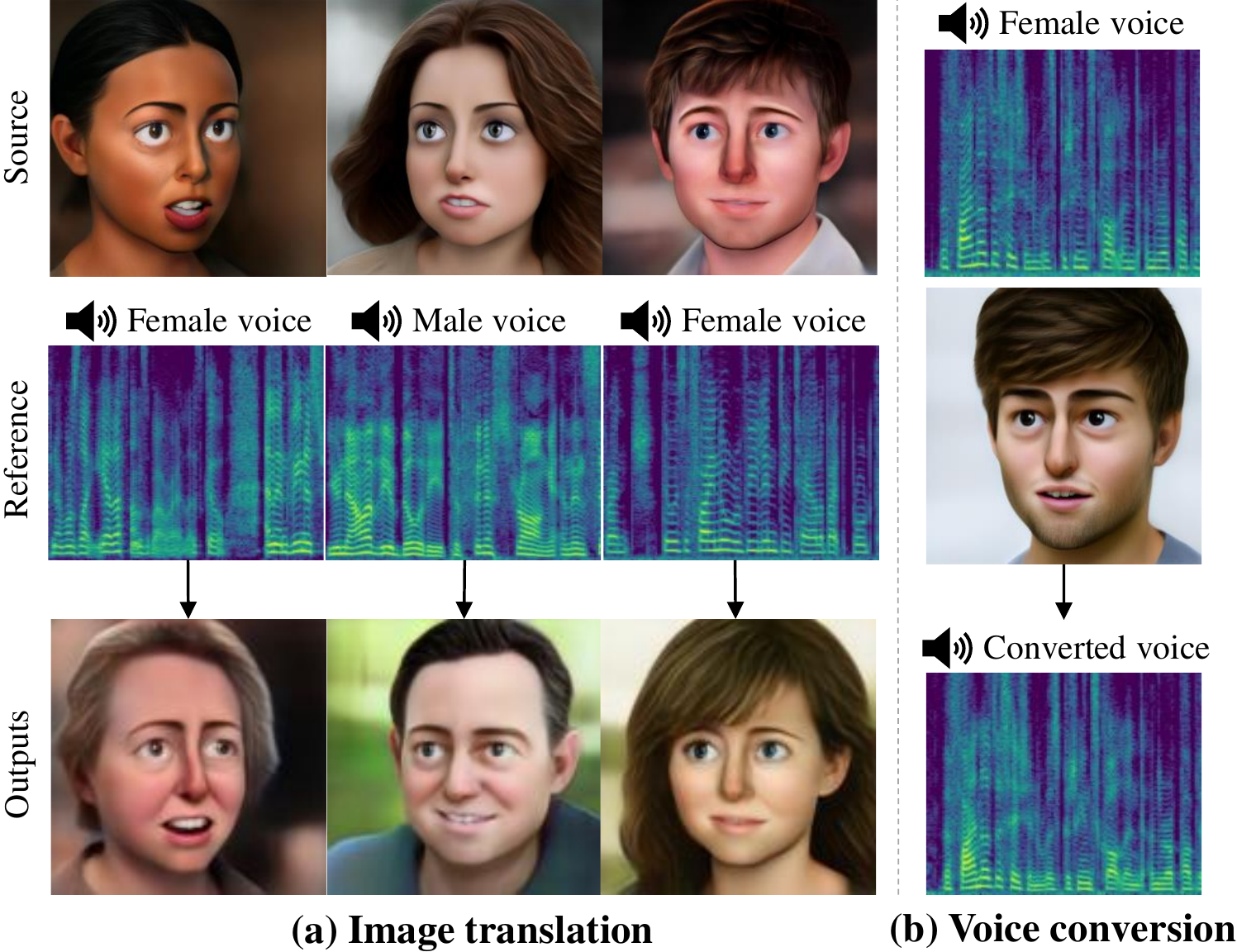}
  \caption{Results of (a) audio-guided image translation and (b) image-guided voice conversion on stylized image data.}
  \label{fig:avatar}
\end{figure}

\subsection{Application}
One of applications of our model is avatar image generation based on voice. For example, users can choose an avatar image from a few preset characters and personalize it using their voice as a reference of the audio-guided image translation mode of the proposed model trained on avatar images. In \Cref{fig:avatar}a, we show the results of the proposed model trained on the stylized images using JoJoGAN \cite{Chong2022JoJoGAN}.
Another application is to convert user's voice to match a user-selected avatar image, as illustrated in \Cref{fig:avatar}b.

\section{Discussion and Conclusion}
We proposed a unified image translation and voice conversion framework with audio and image guidance that converts a face image and voice to plausibly match a user-provided reference in other modalities. Experimental results indicate that our model outperforms baselines in the audio-guided image translation and image-guided voice conversion tasks in terms of quality, image diversity, and face--voice correspondence.  One limitation of the proposed method can be a bias in the training data. For example, the dataset used in this study does not contain many faces and voices of children. Hence, our model still produces faces or voices of adults when we provide faces or voices of a child as a reference. Increasing the diversity of the dataset is for our future work.

\appendix

\section{Training details}
\label{sec:train}
We train our model for 20,000 steps with a batch size of 32. 
The training time is about 2.5 days on a single NVIDIA RTX A6000 GPU with our implementation in PyTorch \cite{Paszke2019Pytorch}.
We set $\lambda_{sty}=1$, $\lambda_{ds}=2$, $\lambda_{cyc}=1$, $\lambda_{asr}=20$, $\lambda_{F0}=5$, $\lambda_{norm}=1$, $\lambda_{advcls}=0.05$, $\lambda_{csc}=1$, $\lambda_{cls}=0.05$, and $\tau=1$.
We use Adam \cite{Kingma2015Adam} optimizer with $\beta_1=0$ and $\beta_2=0.99$ for the image generator $G^{im}$, image style encoder $E^{im}$, image discriminator $D^{im}$, and mapping network $F$, as done in \cite{Choi20StarGANv2}, while we use AdamW for the rest of the audio modules, as done in \cite{Li21StarGANv2VC}. 
The learning rates for $F$ is set to $10^{-6}$, while that of other modules are set to $10^{-4}$.
The weights of all modules are initialized using He initialization \cite{He2015HeInit} and set the biases of $G^{im},E^{im}$, and $D^{im}$ to zero, except for the biases associated with the scaling vectors of AdaIN that are set to one. 
We employ exponential moving averages over parameters \cite{Yazici2019EMA,Karras2018PGGAN} for $G^{au},G^{im},E^{au},E^{im}$, and $F$.
During training, we randomly crop 2.47 s audio, which corresponds to 192 frames in mel-spectrogram, from randomly sampled audio clips. For inference, we use entire audio clip (from 6 to 9 s), as our audio generator and style encoders are fully convolutional networks and can accept an arbitrary length of audio.

\section{Evaluation protocol}
\label{sec:evalprotocol}
This section provides details for the evaluation metrics
and evaluation protocols used in all experiments. We follow the evaluation protocol used in StarGANv2 \cite{Choi20StarGANv2} for the evaluation of the image quality and diversity. \vspace{3mm}\\
\textbf{Frech\'{e}t inception distance (FID)} \cite{Heusel2017FID}\hspace{2mm}
 measures the discrepancy between two groups of images. We use the outputs of the final average pooling layer of the Inception-V3 \cite{Szegedy2016InceptionV3} pretrained on ImageNet dataset. We translate each test image from a source domain into a target domain using 10 reference samples (i.e. 10 images for the image-guided task and 10 utterances for the audio-guided task)
randomly sampled from the test set of a target domain and reference modality. Then, we calculate FID between the translated images and training images in the target domain. We calculate the FID values for every pair of gender domains (i.e. female $\rightarrow$ male and male $\rightarrow$ female ) and report the average value. \vspace{2mm}\\
\textbf{Learned perceptual image patch similarity (LPIPS)} \cite{Zhang2018LPIPS}
measures the diversity of generated images using the L1
distance between features extracted using the AlexNet \cite{Krizhevsky2012AlevNet} pretrained on ImageNet dataset. For each test image from a source domain, we generate 10 outputs of a target domain using
10 randomly sampled reference images or utterances. We then calculate the
average of the pairwise distances among all outputs generated from the same source image (i.e. 45 pairs). We report the average of the LPIPS values over all test images. \vspace{2mm}\\
\textbf{NISQA} \cite{Mittag2020NISQATTS,Mittag2021NISQA} is a neural network-based speech quality prediction model that estimates the mean opinion score of human evaluations on the naturalness of synthesized speech. The highest score is five and the lowest score is one. For audio-guided voice conversion, we generate 140 voices using randomly sampled source and reference audio pairs and compute the NISQA scores using the model provided on the official website\footnote{\url{https://github.com/gabrielmittag/NISQA}}. For image-guided voice conversion, we generate 180 voices using randomly sampled source audio and reference images. Finally, we compute the average of NISQA scores.
\vspace{2mm}\\
\textbf{Speaker similarity (SpkSim)} is the cosine similarity of d-vectors \cite{Variani2014dVector} extracted from a speaker verification model\footnote{\url{https://github.com/resemble-ai/Resemblyzer}}. For audio-guided voice conversion, we compute the SpkSim scores for each converted voice by comparing the converted voice with the reference voice used for the conversion and report the average value. For the evaluation of image-guided voice conversion on GRID, we compare each converted voice with 10 random utterances of the target speaker used for the reference image, and average the scores.
\vspace{2mm}\\
\textbf{Word error rate (WER)} measures the intelligibility of converted voice using automatice speech recognition (ASR). We use a joint CTC-attention based end-to-end ASR system \cite{Kim17HybridASR} provided in ESPNet toolkit \cite{espnet}.  

\begin{table*}
    \centering
    \small
    \caption{Ablation study on image- and audio-guided image translation tasks. Results on GRID and CelebA-HQ(+VCTK) are averaged.}
    \label{tab:ablIT}
    \begin{tabular}{l|cc|cc}
        \toprule
        \multirow{2}{*}[-1pt]{Model} &\multicolumn{2}{c|}{Image-guided} &\multicolumn{2}{c}{Audio-guided}\\
        &FID [$\downarrow$] &LPIPS [$\uparrow$]&FID [$\downarrow$] &LPIPS [$\uparrow$] \\
        \midrule
        StarGANv2 \cite{Choi20StarGANv2}	&54.9	&\textbf{0.200}	&-	&-\\
        $+$ Joint training &52.8	&0.056	&107.2	&0.097\\
        $+$ $\mathcal{L}_{csc}$&	49.3&	0.048&	53.7&	0.049\\
        $+$ Dual domain (=XFaVoT)&	\textbf{34.4}&	0.175&	\textbf{33.3}&	\textbf{0.156}\\
        $-$ $\mathcal{L}_{csc}$&47.0	&0.057&	38.5&	0.036\\
        \bottomrule
    \end{tabular}
\end{table*}

\begin{table*}
    \centering
    \small
    \caption{Ablation study on audio- and image-guided voice conversion tasks on GRID.}
    \label{tab:ablVC}
    \begin{tabular}{l|ccc|ccc}
        \toprule
        \multirow{2}{*}[-1pt]{Model} &\multicolumn{3}{c|}{Audio-guided} &\multicolumn{3}{c}{Image-guided}\\
        &NISQA [$\uparrow$] &SpkSim [$\uparrow$]&WER [$\downarrow$] &NISQA [$\uparrow$] &SpkSim [$\uparrow$] &WER [$\downarrow$] \\
        \midrule
        StarGANv2VC \cite{Li21StarGANv2VC}+one-shot	&4.62	&0.64	&0.448	&- &-&-\\
        $+$ Joint training &4.61	&0.64	&\textbf{0.428}&	4.64&	0.58&	\textbf{0.44}\\
        $+$ $\mathcal{L}_{csc}$&4.62&	0.64&	0.430&	\textbf{4.69}&	0.58&	\textbf{0.44}\\
        $+$ Dual domain ($=$XFaVoT)&\textbf{4.65}&	\textbf{0.75}&	0.455&	4.58&	\textbf{0.66}&	0.48\\
        \bottomrule
    \end{tabular}
\end{table*}

\section{Ablation study }
\label{sec:ablation}

We evaluate individual components we newly introduced in XFaVoT. For image translation tasks, we start from our base model, StarGANv2 \cite{Choi20StarGANv2}, and cumulatively add each component. \Cref{tab:ablIT} shows the FID and LPIPS for several configurations. When we add audio modules with the same domain codes as image modules (i.e. gender domain) and jointly train the model for the four tasks without the cross-modality style consistency loss $\mathcal{L}_{csc}$, we obtain the slight improvement on FID with image guidance. However, we observe a mode collapse for image-guided image translation and obtained low LPIPS. Interestingly, we do not observe the severe mode collapse with audio guidance and obtain higher LPIPS than the image-guided task. However, FID becomes high with audio guidance, indicating low image quality. By adding the $\mathcal{L}_{csc}$, we observe improvements on FID and degradation on LPIPS for both image- and audio-guided tasks. When we introduce the proposed dual domain types (gender and speaker identity) for the discriminators and mapping network, the model becomes our proposed XFaVoT, and FID and LPIPS are significantly improved for both image- and audio-guided tasks. Finally, we also evaluate the XFaVoT trained without the $\mathcal{L}_{csc}$ and observe that FIDs and LPIPSs are significantly degraded. This results indicate that both $\mathcal{L}_{csc}$ and dual domain codes are necessary for achieving high quality and diversity for the cross- and intra-modal image translation tasks. 

For voice conversion tasks, we start from a modified version of StarGANv2-VC \cite{Li21StarGANv2VC}, where we extend StarGANv2-VC to one-shot voice conversion by introducing gender-domain specific heads for the style encoder, mapping network, and discriminators. Similar to the evaluation of the image translation tasks, we then add the join training of image modules, the $\mathcal{L}_{csc}$, and the dual domain codes, cumulatively. The results are summarized in \Cref{tab:ablVC}. Overall, we observe similar NISQAs and WERs for all configurations, however, SpkSim scores are clearly improved when we introduce the dual domain codes. The results indicates that using the dual domain code is essential for achieving high speaker similarity with the reference while enabling the one-shot voice conversion.

\begin{figure}[t]
  \centering
  \includegraphics[width=\linewidth]{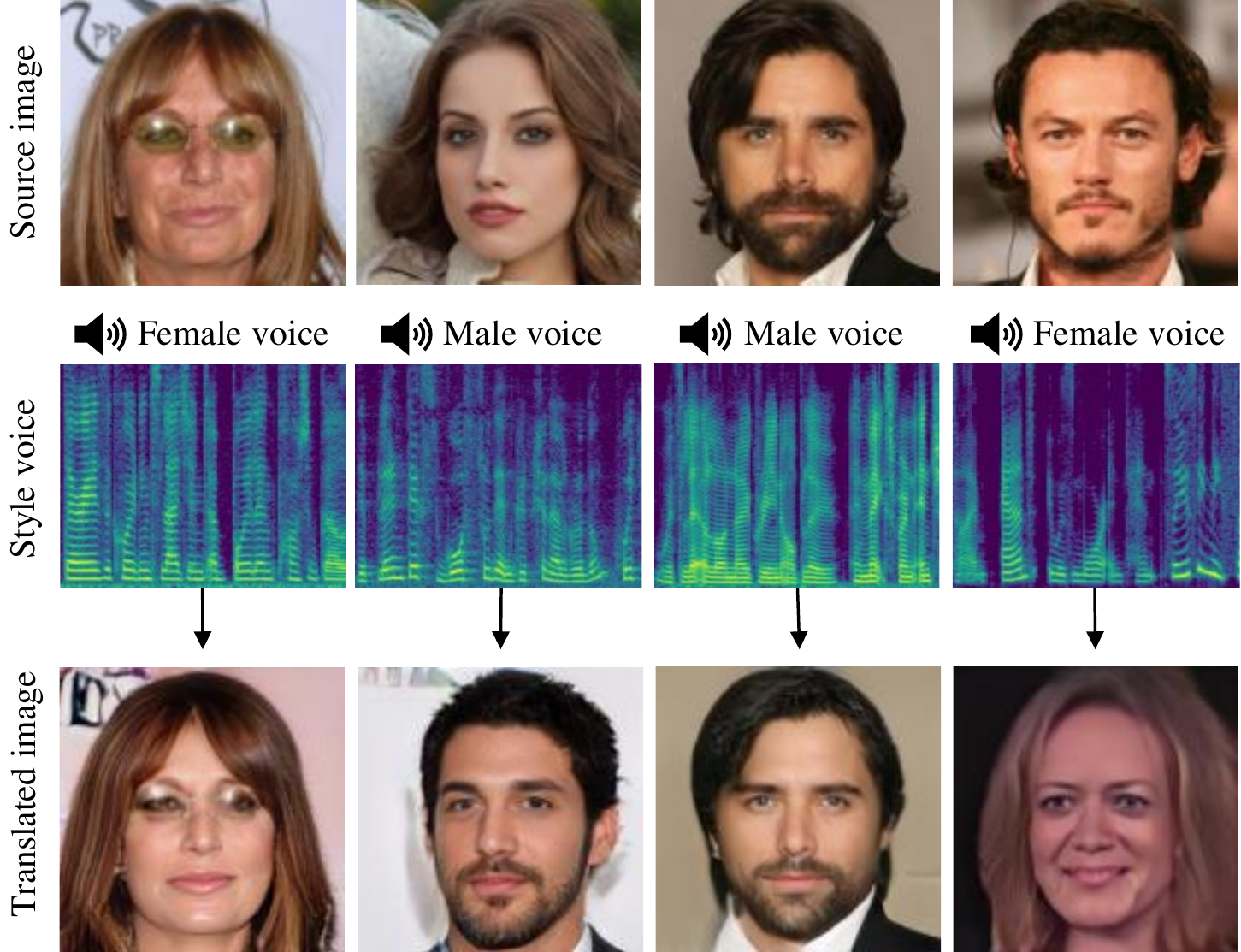}
  \caption{Additional results of audio-guided image translation.}
  \label{fig:addresit}
\end{figure}

\begin{figure}[t]
  \centering
  \includegraphics[width=\linewidth]{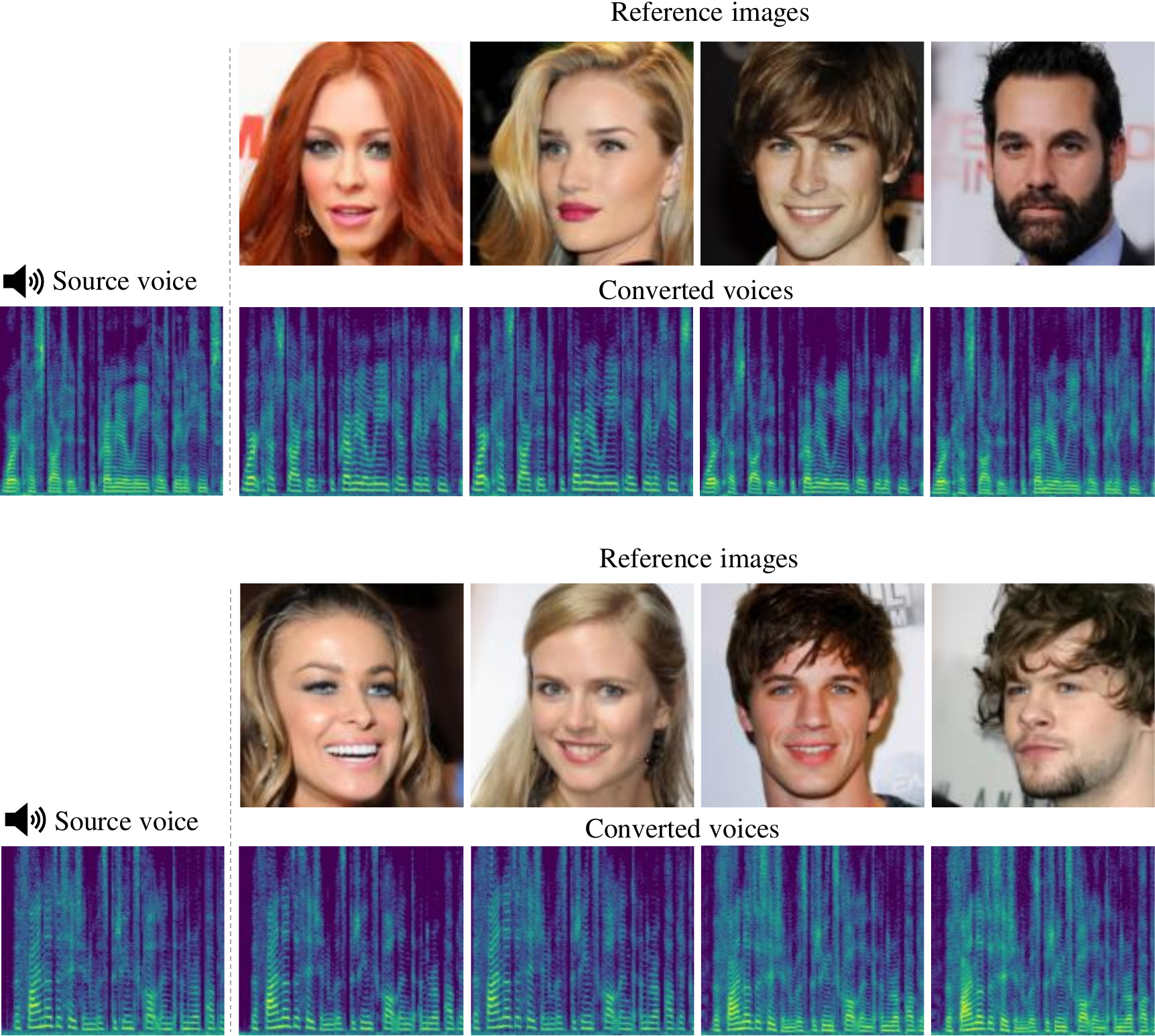}
  \caption{Additional results of image-guided voice conversion.}
  \label{fig:addresvc}
\end{figure}

\section{Additional results}
We provide additional results of audio-guided image translation and image-guided voice conversion in \Cref{fig:addresit} and \Cref{fig:addresvc}, respectively. 

\section{Visualization of learned style embedding space}
\begin{figure}[t]
  \centering
  \includegraphics[width=\linewidth]{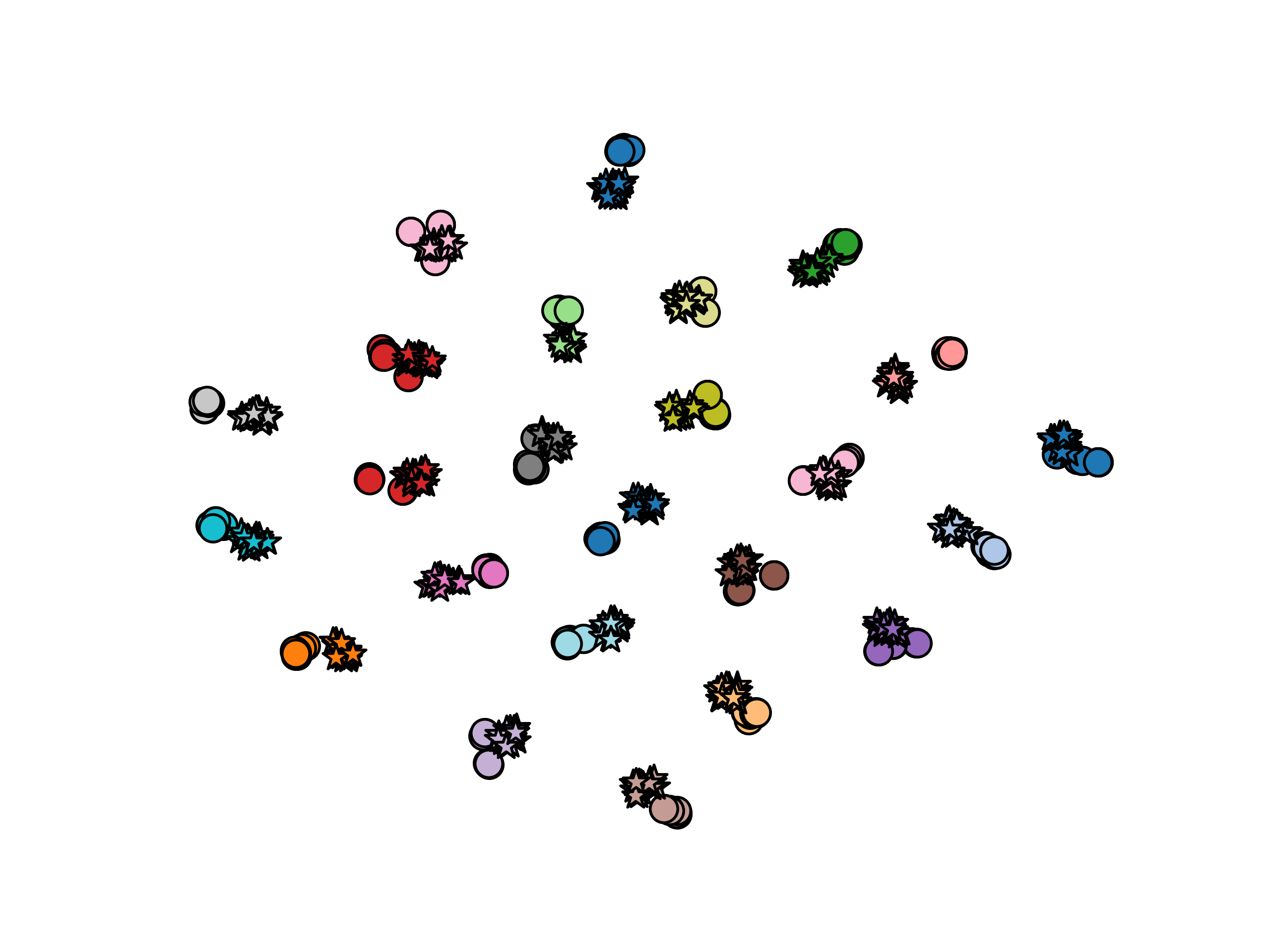}
  \caption{Visualization of style vectors using t-SNE. $\circ$ and $\star$ denote style vectors extracted from image and audio of GRID, respectively. The colors represent speaker identities.}
  \label{fig:tSNE}
\end{figure}

We visualize the learned style embedding space using t-SNE \cite{Maaten2008tSNE}. In \Cref{fig:tSNE}, we plot the style vectors extracted from the faces and voices of the GRID speakers using image and audio style encoders, respectively. As observed, style vectors extracted from faces (denoted with $\circ$) and voices (denoted with  $\star$) of the same speaker (denoted with colors) are placed closer than that of other speakers. This results indicate that the audio and image style encoders successfully learn a common style space for audio and image that capture speaker specific styles. 

\section{AdaIN-CVC baseline}
\label{sec:baseline}
We provide the details of the AdaIN-CVC baseline. The model architecture is based on AdaIN-VC \cite{chou2019adainvc}, and the speaker encoder is replaced with the face encoder used in CVC \cite{Kameoka2019CMVC}. We base our code on the official implementation of AdaIN-VC\footnote{\url{https://github.com/jjery2243542/adaptive_voice_conversion}}. We train the model by following the official implementation using the same audio-visual datasets as the proposed model, namely, GRID, LRS3, and Lip2Wav. Note that the training framework require audio-image pair data and cannot use audio-only or image-only data.


\section{Network architecture}
We base our network architecture on StarGANv2 \cite{Choi20StarGANv2,Li21StarGANv2VC}.\vspace{2mm}\\
\textbf{Generators} (\Cref{tab:gi},\Cref{tab:ga})\textbf{.} \hspace{1mm} The image and audio generators consist of four downsampling blocks, four intermediate blocks, and four upsampling blocks, all of which consist of preactivation residual blocks \cite{He2106ResNet}.
The instance normalization (IN) \cite{Ulyanov2016IN} is used for downsampling blocks, while the adaptive instance normalization (AdaIN) \cite{Huang2017AdaIN} is used for up-sampling blocks. A style code is fed to all AdaIN layers to provide scaling and shifting vectors through learned affine transformations. We remove skip connections with the adaptive wing based heatmap \cite{Wang2019Heatmap} for upsampling layers of the image generator. We concatenate F0 features extracted from the source utterance using a pre-trained joint detection and classification (JDC) F0 extraction network \cite{Kum2019JDC} at the midle of the intermediate blocks, as done in \cite{Li21StarGANv2VC}.
\vspace{2mm}\\
\textbf{Mapping network} \hspace{1mm} consists of an MLP with $K$ output branches, where $K$ denotes the number of total domains (2 gender $+$ 128 speaker identity = 130). The MLP consists of shared four fully connected layers, followed by four domain-specific fully connected layers for each domain. The dimensions of the latent code, the hidden layer, and the style vector are set to 16, 512, and 64, respectively. The latent code is sampled from the standard Gaussian distribution. 
\vspace{2mm}\\
\textbf{Style encoders} (\Cref{tab:ei}, \Cref{tab:ea})\textbf{.} \hspace{1mm} Our style encoders consist of a CNN with $K_g$ output heads, where $K_g$ is the number of gender domains (i.e.~$K_g=2$). Five and four pre-activation residual blocks are shared among all domains in image- and audio-style encoders, respectively, followed by one gender-domain specific fully connected layer.  The output dimension of the domain-specific heads is 64, which is the dimension of the style vector.
\vspace{2mm}\\
\textbf{Discriminators} (\Cref{tab:ei}, \Cref{tab:ea})\textbf{.}\hspace{1mm} Multi-task discriminators \cite{Mescheder2018}, which contain multiple linear output branches, are used for the image and audio discriminators. Five and four pre-activation residual blocks with leaky ReLU are used for the image and audio discriminators, respectively. The image discriminator has $K_g$ fully-connected layers for real/fake classification of each gender domain, while the audio discriminator use $K=K_g+K_y$ fully-connected layers for real/fake classification of each gender domain and speaker identity domain, where $K_y=128$ indicates the number of speaker identity domains. The audio classification network has the same network architecture as the audio discriminator, where the $K=K_g+K_y$ output branches are treated as a $K_g$-class and $K_y$-class classification heads for source gender and speaker identity classifications, respectively.

\begin{table}
    \centering
    \small
    \caption{Image generator architecture.}
    \label{tab:gi}
    \begin{tabular}{cccc}
        \toprule
        Layer &Resample &Norm. &Output shape\\
        \midrule
        Input & - & - 	&128$\times$128$\times$3 \\
        Conv~1$\times$1 &-&-&128$\times$128$\times$128 \\
        ResBlock &AvgPool&IN&64$\times$64$\times$256 \\
        ResBlock &AvgPool&IN&32$\times$32$\times$512 \\
        ResBlock &AvgPool&IN&16$\times$16$\times$512 \\
        ResBlock &AvgPool&IN&8$\times$8$\times$512 \\
        \midrule
        ResBlock &-&IN&8$\times$8$\times$512 \\
        ResBlock &-&IN&8$\times$8$\times$512 \\
        ResBlock &-&AdaIN&8$\times$8$\times$512 \\
        ResBlock &-&AdaIN&8$\times$8$\times$512 \\   
        \midrule
        ResBlock &Upsample&AdaIN&16$\times$16$\times$512 \\ 
        ResBlock &Upsample&AdaIN&32$\times$32$\times$512 \\ 
        ResBlock &Upsample&AdaIN&64$\times$64$\times$256 \\   
        ResBlock &Upsample&AdaIN&128$\times$128$\times$128\\ 
        Conv~1$\times$1 &-&-&128$\times$128$\times$3 \\
        \bottomrule
    \end{tabular}
\end{table}

\begin{table}
    \centering
    \small
    \caption{Audio generator architecture.}
    \label{tab:ga}
    \begin{tabular}{cccc}
        \toprule
        Layer &Resample &Norm. &Output shape\\
        \midrule
        Input & - & - 	&80$\times$192$\times$1 \\
        Conv~1$\times$1 &-&-&80$\times$192$\times$64 \\
        ResBlock &AvgPool&IN&40$\times$96$\times$128 \\
        ResBlock &AvgPool&IN&20$\times$96$\times$256 \\
        ResBlock &AvgPool&IN&10$\times$48$\times$512 \\
        ResBlock &AvgPool&IN&5$\times$48$\times$512 \\
        \midrule
        ResBlock &-&IN&5$\times$48$\times$512 \\
        ResBlock &-&IN&5$\times$48$\times$512 \\
        Concat. &-&-&5$\times$48$\times$640 \\
        ResBlock &-&AdaIN&5$\times$48$\times$640 \\
        ResBlock &-&AdaIN&5$\times$48$\times$640 \\   
        \midrule
        ResBlock &Upsample&AdaIN&10$\times$48$\times$512 \\ 
        ResBlock &Upsample&AdaIN&20$\times$96$\times$256 \\   
        ResBlock &Upsample&AdaIN&40$\times$96$\times$128\\ 
        ResBlock &Upsample&AdaIN&80$\times$192$\times$64 \\ 
        Conv~1$\times$1 &-&-&80$\times$192$\times$1 \\
        \bottomrule
    \end{tabular}
\end{table}
\begin{table}
    \centering
    \small
    \caption{Image style encoder and discriminator architectures. $d$ and $k$ represent the output dimension and number of domains, respectively. We use $d=64,k=K_g$ for style encoder and $d=1,k=K_g$ for discriminator.}
    \label{tab:ei}
    \begin{tabular}{cccc}
        \toprule
        Layer &Resample &Norm. &Output shape\\
        \midrule
        Input & - & - 	&128$\times$128$\times$3 \\
        Conv~1$\times$1 &-&-&128$\times$128$\times$128 \\
        ResBlock &AvgPool&IN&64$\times$64$\times$256 \\
        ResBlock &AvgPool&IN&32$\times$32$\times$512 \\
        ResBlock &AvgPool&IN&16$\times$16$\times$512 \\
        ResBlock &AvgPool&IN&8$\times$8$\times$512 \\
        ResBlock &AvgPool&IN&4$\times$4$\times$512 \\
        \midrule
        LReLU &-&-& 4$\times$4$\times$512 \\
        Conv~4$\times$4 &-&-&1$\times$1$\times$512 \\
        LReLU&-&-&1$\times$1$\times$512 \\
        Linear$\times k$ &-&-& $d\times k$\\
        \bottomrule
    \end{tabular}
\end{table}

\begin{table}
    \centering
    \small
    \caption{Audio style encoder, discriminator, and classifier architectures. $d$ and $k$ represent the output dimension and number of domains, respectively. We use $(d,k)=(64,K_g), (1, K), (1, K)$ for style encoder, discriminator, and classifier, respectively.}
    \label{tab:ea}
    \begin{tabular}{cccc}
        \toprule
        Layer &Resample &Norm. &Output shape\\
        \midrule
        Input & - & - 	&80$\times$192$\times$1 \\
        Conv~1$\times$1 &-&-&80$\times$192$\times$64 \\
        ResBlock &AvgPool&IN&40$\times$96$\times$128 \\
        ResBlock &AvgPool&IN&20$\times$48$\times$256 \\
        ResBlock &AvgPool&IN&10$\times$24$\times$512 \\
        ResBlock &AvgPool&IN&5$\times$12$\times$512 \\
        \midrule
        LReLU &-&-& 5$\times$12$\times$512 \\
        Conv~5$\times$5 &AvgPool&-&1$\times$1$\times$512 \\
        LReLU&-&-&1$\times$1$\times$512 \\
        Linear$\times k$ &-&-& $d\times k$\\
        \bottomrule
    \end{tabular}
\end{table}

{\small
\bibliographystyle{ieee_fullname}
\bibliography{cv,vc}
}

\end{document}